\documentclass[fleqn,10pt]{wlscirep}
\usepackage[utf8]{inputenc}
\usepackage[T1]{fontenc}
\usepackage{amssymb}
\usepackage{amsmath}
\usepackage{lineno}
\usepackage{bm}
\usepackage{siunitx}
\usepackage{csquotes}
\usepackage{subfig}
\usepackage{cleveref}

\usepackage{todonotes}

\title{What is the effect of running-specific prostheses on long jumps? Optimization-based prediction and analysis using biomechanical models}

\author[1]{Anna Lena Emonds}
\author[2]{Johannes Funken}
\author[2]{Wolfgang Potthast}
\author[3,4,*]{Katja Mombaur}
\affil[1]{Institute of Computer Engineering, Heidelberg University, 69120 Heidelberg, Germany}
\affil[2]{Institute of Biomechanics and Orthopaedics, German Sport University Cologne, 50933 Köln, Germany}
\affil[3]{Institute for Anthropomatics and Robotics, Optimization and Biomechanics for Human-Centred Robotics, Karlsruhe Institute of Technology, 76131 Karlsruhe, Germany}
\affil[4]{Canada Excellence Research Chair in Human-Centred Robotics and Machine Intelligence, Systems Design Engineering \& Mechanical and Mechatronics Engineering, University of Waterloo, Waterloo, N2L 3G1, Canada}

\affil[*]{katja.mombaur@kit.edu}

\begin{abstract}
Long jumpers with below the knee amputation (BKA) that take off from their running-specific prosthesis (RSP) improved performances significantly over the last years. The long jump biomechanics differs compared to athletes without BKA and the question arises whether the spring-like properties of the RSP facilitate achieving long jumping distances. The aim of this work is to propose a long jump model for athletes with and without BKA, to evaluate it and to apply it for comparing long jump motions with and without RSP.  
We establish rigid multi-body system models of one athlete with and one athlete without below the knee amputation (BKA). Long jump motions are computed by solving a specific optimal control problem (OCP) with constraints enforcing a physically correct dynamics, both for motion reconstruction or motion synthesis. 
With the proposed long jump model, we are able to compute realistic long jump motions. We discuss the causes of differences in measured long jumps and show directions for eliminating them. For both athletes, the synthesized solutions reveal potential for performance improvement. The jumping distance of the athlete without BKA is \SI{64}{\cm} (\SI{6.9}{\percent}) longer than the one of the athlete with BKA in the synthesized solutions.
\end{abstract}
\begin{document}

\flushbottom
\maketitle
%
%
\thispagestyle{empty}

\section*{Introduction}
\label{sec:introduction}
Understanding human motions and the underlying biomechanics is a main focus of research in various disciplines, e.g. computer animation, medicine, sport science or robotics. A comprehensive knowledge of the principles of a specific motion is useful in the design process of prosthesis and assistive devices and in improving therapy or performance. Concretely applied to the long jump, a precise understanding of the movement allows the identification of potential for improvement in a given athlete, the calculation of a maximum possible jumping distance, and also the comparison of the performances of athletes with and without a running-specific prostheses (RSP).

For non-amputee long jumpers, detailed analyses of the biomechanics of run-up, take-off and landing have already been conducted (e.g. \cite{Alexander1990,Arampatzis1999,Hay1986b,Hay1993,Lees1993,Lees1994,Nixdorf1990,Linthorne2008}). 
All three components influence long jump performance: The athlete must use the run-up to achieve the optimal horizontal run-up velocity. Towards the end of the run-up he must get himself in a good position for placing the take-off foot on the board and for the subsequent take-off. 
During the take-off step, the athlete must generate a high vertical velocity without loosing too much horizontal velocity, and at the same time hit the optimal take-off angle of the center of mass (CoM).
Successful athletes typically use a technique called \enquote{pivoting}: the body's CoM is rotated around the fixed ground contact point, using the take-off leg as a rigid lever arm \cite{Lees1994}. 
Finally, the landing technique is of crucial importance, for example to prevent falling backwards after landing with the forward stretched legs.

Studies have also examined the long jump motions of athletes with below the knee amputation (BKA), particularly in comparison to the motions of athletes without BKA (e.g., \cite{Nolan2012,Willwacher2017b},\cite{Funken2019a},\cite{Funken2019c}). 
Willwacher and colleagues \cite{Willwacher2017b} compared three long jumpers with unilateral BKA to a group of seven non-amputee athletes and concluded that the athletes with BKA were able to jump off more efficiently, but run up more slowly. Although the authors showed that the prosthesis can store energy more efficiently than the biological foot, no definitive statement could be made regarding the overall long jump performance of athletes with BKA compared to non-amputee athletes. With the same experimental setup, Funken et al. \cite{Funken2019c} reported that athletes with BKA move mainly in the sagittal plane during the take-off step (with the prosthetic device). Furthermore, amputee athletes stiffen their knee during the take-off step to employ the prosthesis as an efficient spring \cite{Funken2019a}. Ashcraft and Grabowski \cite{Ashcraft2026} studied the effect of the prosthesis' and the take-off board's stiffness on the long jump performance. Their findings suggest that the difference between athletes with and without BKA in maximum run-up velocity and jump distance are small.

However, what various authors \cite{McGowan2012,Weyand2010} have noted in the debate regarding sprinting with RSP applies to the long jump too: General statements are difficult if not impossible to draw from measurements, since there are only few athletes with BKA performing at world level. In addition, it is not possible to take measurements of an amputee athlete with and without BKA in order to compare him to himself.

Here, computer simulations can help to get closer to an answer to the question of advantage and disadvantage due to the use of RSP. Indeed, in a computer simulation, it is possible to compare the model of an amputee athlete with a non-amputee model version of oneself. For example, Emonds and colleagues have compared sprinting motions with and without RSP \cite{Emonds2019a,Emonds2019b}. Hase et al. \cite{Hase2017} have proposed an computer simulation model  for improving the long jump performance of amputee athletes.

Hence, the overarching problem is to understand the effect of a RSP on long jump motions in order to be able to discuss advantages and disadvantages resulting from the RSP and to point directions for performance improvement. 
The aim of this work is to propose a long jump model based on subject-specific models of athletes with and without BKA and optimal control problem formulations which is capable of predicting realistic motions. In this sense, the first purpose of this research is to evaluate the proposed long jump model and the second purpose is to employ it for predicting and comparing the motions of athletes with and without BKA.  
 
Hence, in this research we want to answer the following research question: Is the proposed long jump model valid for generating realistic long jump motions with and without RSP? 
For answering the research question, we compare the model prediction and reality regarding performance indicators like jump distance and characteristic kinematic and kinetic variables like ground reaction forces, joint angles and CoM motion.

\section*{Materials and Methods}
\label{sec:methods}
\begin{figure}[tbp]
\centering
\includegraphics[width=0.75\textwidth]{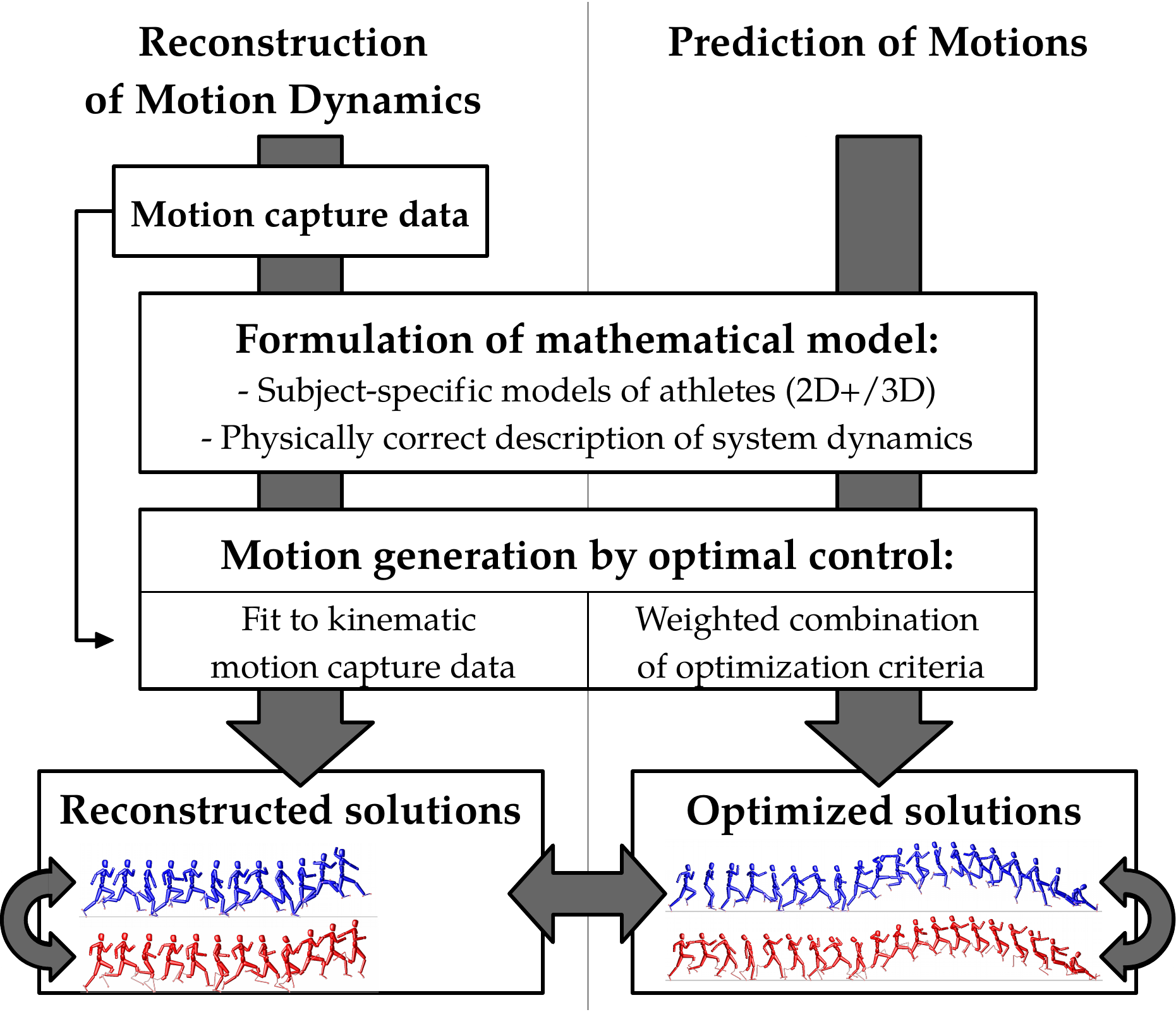}
\caption{Overview of the study design: The computations are divided into two steps, the first reconstructing the motion dynamics based on motion capture recordings and the second predicting long jump motions based on a combination of optimization criteria. The solutions are then used to conduct two distinct comparisons, one between the reconstructed and the optimized solutions and the other between the athletes with and without BKA.}
\label{fig:overview}
\end{figure}

\unskip
To address the objective of this paper, we established the following study design which is schematically shown in Figure \ref{fig:overview}: We formulate a realistic computer model of a long jump motion consisting of a subject-specific modeling of the athletes, a physically correct description of the dynamics, and an optimal control problem to compute the motion. In a first step, we validate the correctness of the models based on kinematic motion capture data by reconstructing the dynamics. In a second step, we predict long jump motions based on a specific combination of optimization criteria to compare athletes with and without BKA. The solutions are then used to conduct two distinct comparisons, one between the reconstructed and the optimized solutions and the other between the athletes with and without BKA.

\subsection*{Modeling of Unilateral Transtibial Amputee and Non-Amputee Long Jump}
\label{subsec:modeling}
\unskip
\subsubsection*{Subject-Specific Models of Athletes}
\label{subsubsec:modelingSubjectSpecific}
\begin{figure}
\centering	
\includegraphics[width=0.6\textwidth]{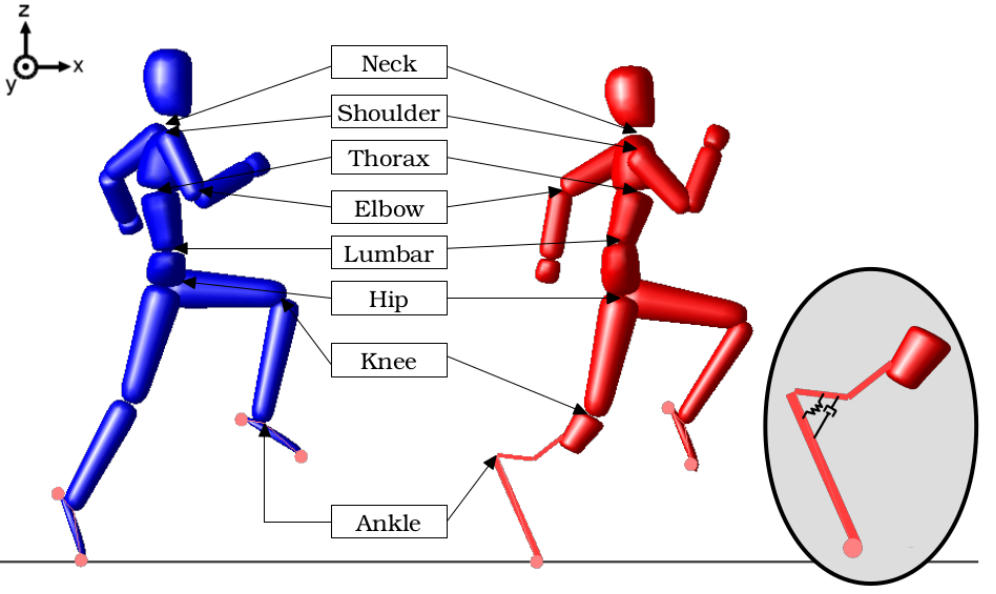}
\caption{Subject-specific models of a non-amputee and an amputee athlete: Visualization and degrees of freedom. The coordinate system is aligned such that the x-axis points in the forward direction of the sprinting motion. Movements to the right and left are along the y-axis, and movements up and down are along the z-axis.}
\label{fig:models}
\end{figure}

We created subject-specific rigid multi-body system models of one non-amputee athlete (\SI{1.83}{\m}, \SI{84.1}{\kg}) and one athlete with unilateral BKA (\SI{1.84}{\m}, \SI{76.0}{\kg}) by extrapolation of the de Leva data \cite{DeLeva1996} to the individual mass and height. The non-amputee model has 16 segments: head, upper and lower arms, hands, three spinal segments, thighs, shanks and feet. The right foot and part of the right shank is replaced by a three-segment model of the prosthetic device for the model of the amputee athlete. A visualization of both models including the model of the RSP is shown in Figure \ref{fig:models}. For the first step of the computations, the reconstruction of the motion dynamics from purely kinematic data, we considered a three-dimensional model with 31 (non-amputee model)/30 (amputee model) degrees of freedom (DOFs) that we will refer to as 3D model in the following. Six global DOFs describe the overall motion of the system with the pelvis segment as floating base. The remaining DOFs are internal rotations of the joints. There is one DOF less in the case of the amputee athlete as we only consider rotations in the sagittal plane for the joint in the RSP. For the second step of the computations, we use simplified models that are basically restricted to the sagittal plane, except for the shoulder joints (resulting in 20 DOFs for both models), that we will refer to as 2D\texttt{+} model in the following. The wrists are fixed for both the 2D\texttt{+} and 3D models.  All internal joints are driven by joint torque actuators which approximate the action of all related muscles at a joint. A RSP is a passive device, i.e. not actuated. Made from compliant materials, it shows an overall deformation such that it does not have a specific \enquote*{ankle joint}. Instead we use the point of greatest curvature which is likewise the most posterior point of the RSP\cite{Willwacher2017b} to define the \enquote*{ankle joint} at the connection of the two lower segments. The joint between the two upper segments and the attachment of the prosthesis model to the residual limb are fixed. At the so-defined prosthetic ankle joint a passive torque $\tau_{F}$ is generated:      

\begin{equation}
  \tau_{F} \! \left( q, \dot{q} \right) = - d \dot{q} - k \left( q - \vartheta_{0} \right) \; .
  \label{eq:springDamper}
\end{equation}

It is computed based on the spring constant $k$, damping constant $d$ and the rest position $\vartheta_0$, current angle $q$ and current angular velocity $\dot{q}$ of the RSP. 

\subsubsection*{Mathematical Description of Long Jump Motions}
\label{subsubsec:modelingMathematical}
\begin{figure}
\centering
\includegraphics[width=\textwidth]{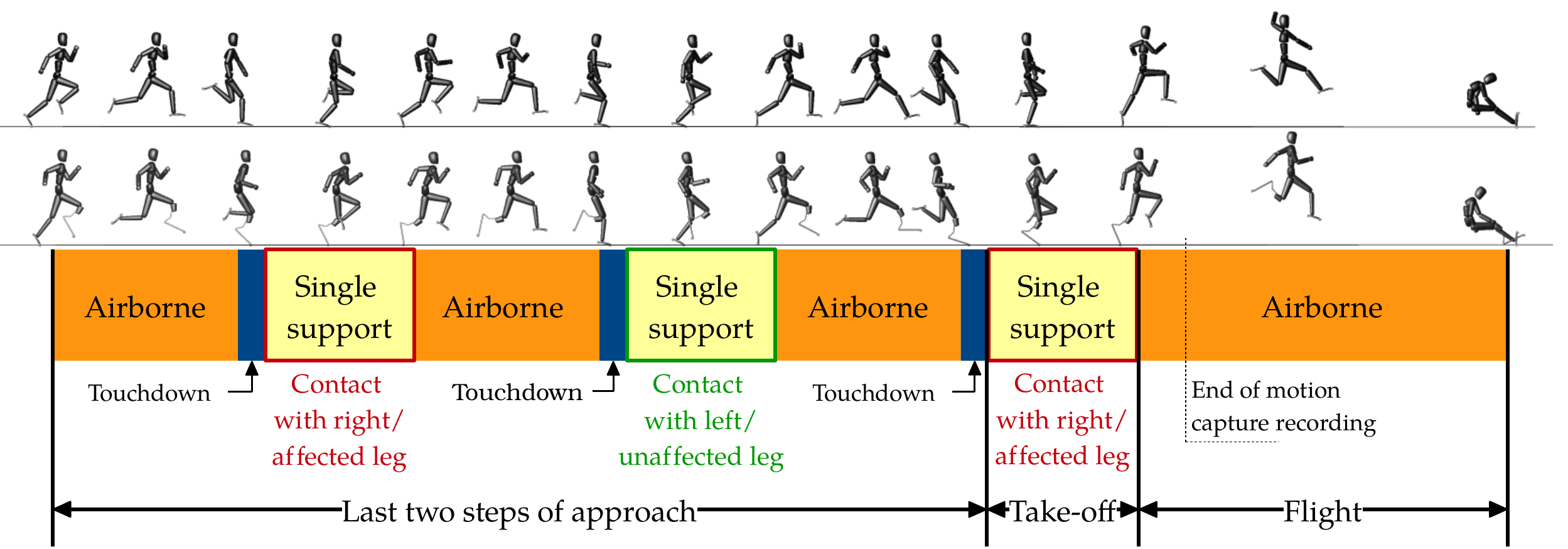}
\caption{Phase description of the investigated long jump motions for a non-amputee athlete (top) and an athlete with below the knee amputation (bottom)}
\label{fig:phasesLongJump}
\end{figure}

To formalize the long jump motion mathematically, we divide the motion into phases. Here, we follow Hay's division into four different parts: approach, takeoff, flight and landing \cite{Hay1986b}. Specifically, we model the last two steps of the approach, the take-off and the flight (see Figure \ref{fig:phasesLongJump}).   
For the concrete modeling, we assume that the motion is essentially described by two states: Either the athlete has no contact with the ground (airborne or flight phase) or there is unilateral point ground contact with the ball of the foot (contact phase). These two kinds of phases form an alternating sequence and are connected by lift-off and touchdown events. The latter are modeled as transition phases of zero length due to the completely inelastic touchdown resulting in velocity discontinuities. Hence, we model the investigated motion by a total of ten phases. Each of them is governed by its own set of ordinary differential equations (ODEs) or differential-algebraic equations (DAEs).

Here, the equations of motion depend on the current state of the system. We describe this by generalized coordinates. For example, the vector of generalized positions combines the variables describing the spatial position and orientation of the athlete and the variables describing the internal joint angles.

Within the flight phases, no ground contact occurs, and hence the equation of motion is

\begin{equation}
  M \! \left( q \right) \ddot{q} + N \! \left( q, \dot{q} \right) = \tau \; ,
  \label{eq:modelingODEFlight}
\end{equation}

with the generalized positions, velocities, accelerations and forces $q$, $\dot{q}$, $\ddot{q}$ and $\tau$. The inertial properties of the system enter the equation of motion via the positive definite matrix $M$. All external forces (gravity, air friction, muscle torques, torques of spring-damper system) are summarized in the generalized forces vector $\tau$, whereas all non-linear effects (e.g. centrifugal or Coriolis forces) are described by the vector $N$. 

The single support contact is modeled by a non-sliding and rigid contact between the hallux point or tip of the RSP, respectively, and the ground. Hence, the number of DOFs of the system is reduced by two ($x$- and $z$-coordinates of the contact point). To make the formulation within the optimal control problem easier, we aim to keep the same number of coordinates which is achieved by introduction of $m$ holonomic scleronomic constraints $g: \mathbb{R}^{n_{\text{dof}}} \rightarrow \mathbb{R}^{m}$. The following index-3 DAE system then governs the motion:

\begin{subequations}
  \label{eq:modelingEoMContact}%
\begin{align}
  M \! \left( q \right) \ddot{q} + N \! \left( q, \dot{q} \right) &= \tau + G \! \left( q \right)^{T} \lambda \\
  g \! \left( q \right) &= 0 \; .
\end{align}
\end{subequations}

Here, $G = \left(\partial g/\partial q \right)$ denotes the contact Jacobian and $\lambda \in \mathbb{R}^{m}$ the contact forces. The solubility of the system depends on the non-redundant conditions in $g$. Differentiation yields the linear system

\begin{equation}
    \begin{bmatrix} M & G^{T} \\ G & 0    \end{bmatrix} 
    \begin{bmatrix} \ddot{q}  \\ -\lambda  \end{bmatrix} 
  = \begin{bmatrix} -N + \tau \\ \gamma 	 \end{bmatrix} \; .
  \label{eq:modelingEoMContactLinear}
\end{equation}

with the unknowns $\ddot{q}$ and $\lambda$ and the Hessian matrix $\gamma$ which can be computed by differentiation of the position constraints. During contact, the contact point should be firmly on the ground which is assured by enforcing a positive vertical ground reaction force,  

\begin{equation}
  F_{z}^{*} \! \left( x \! \left( h_{i} \right) , u \! \left( h_{i} \right)\right) \geq 0 \; ,
  \label{eq:constraintsContactPhase}
\end{equation}

with \enquote*{*} denoting the respective contact point. Equivalence of Eq.s \eqref{eq:modelingEoMContact} and \eqref{eq:modelingEoMContactLinear} is given if the invariants of the constraints are fulfilled at the beginning of the contact phase:

\begin{subequations}
\begin{align}
  g_{pos} &= g \! \left( q \! \left( t \right) \right) = 0 \;  \text{and}\\
  g_{vel} &= G \! \left( q \! \left( t \right) \right) \cdot \dot{q} \! \left( t \right) = 0 \; .
  \label{eq:equivalenceInvariants}
\end{align}
\end{subequations}

The beginning of the contact phase is marked by the touchdown event. Touchdown occurs when the vertical position of the respective contact point becomes zero:

\begin{subequations}
\begin{equation}
    P_{z}^{*} \! \left( x \! \left( h_{i+1} \right) \right) x = 0 \; . 
	\label{eq:constraintTouchdownPosition}
\end{equation}

In addition, we require that the contact point is instantaneously rigidly attached to the ground without bouncing of again which is formalized by

\begin{equation}
    -V_{z}^{*} \! \left( x \! \left( h_{i+1} \right) \right) x \geq 0 \; .
    \label{eq:constraintTouchdownVelocity}
\end{equation}
\end{subequations}

This approximation neglects fast timescale effects of the real contact which research has identified to be valid for whole-body model running motion prediction \cite{Schultz2010,Mombaur2014}. The discontinuities in the generalized coordinates on velocity level due to the completely inelastic touchdown result in the generation of a contact impulse $\Lambda$ and are the solution of the system

\begin{equation}
    \begin{bmatrix} M & G^{T} \\ G & 0    \end{bmatrix}
	\begin{bmatrix} v_{+}     \\ \Lambda  \end{bmatrix}
  = \begin{bmatrix} M v_{-}   \\ 0        \end{bmatrix} \; .
  \label{eq:modelingDiscontinuity}
\end{equation}

Here, $v^{-}$ and $v^{+}$ denote the velocities before and after the collision. In contrast to the touchdown, lift-off events are do not result in discontinuities and are therefore not modeled by a distinct phase. A vanishing vertical ground reaction force initiates the lift-off event:

\begin{equation}
  F_{z}^{*} \! \left( x \! \left( h_{i} \right) , u \! \left( h_{i} \right)\right) = 0 \; .
  \label{eq:constraintsFlightPhase}	
\end{equation}

The Rigid Body Dynamics Library (RBDL, \cite{Felis2017}) is used for the computation of the equations of motion. 

\subsection*{Optimal Control Problem Formulation For Motion Generation}
\label{subsec:ocp}
For the computation of realistic long jump motions, we solve an optimal control problem (OCP). The aim of such an OCP is the computation of a set of trajectories (i.e., a description of the desired motion) such that a combination of chosen optimization criteria is minimized while respecting the system dynamics and given constraints.

The following general formulation of an optimal control problem (OCP) is applied for the investigation of long jump motions with and without RSP:

\begin{subequations}
\begin{equation}
 \hspace*{1cm} \min_{x \left( \cdot \right), \, u \left( \cdot \right), \, p}  \Phi \! \left( x \! \left(t\right), u \! \left(t\right), p, h \right) \label{eq:objectiveFunction}
\end{equation}
subject to
\begin{align}
 \dot{x} \! \left( t \right) &= f_{i} \! \left(t, x \! \left(t\right), u \! \left(t\right), p\right) \;,  &t\in\left[h_{i-1},h_{i}\right] \;, \label{eq:ode} \\
 x \!\left(h_{i}^{+}\right) &= c_{i} \! \left( x\!\left(h_{i}^{-}\right), p \right) \;, &i=1,\dots,m \;, \label{eq:switchingCondition} \\
 g\! \left(t, x\! \left(t\right), u\! \left(t\right), p \right) &\geq 0 \;, &t\in\left[h_{i-1},h_{i}\right] \;, \label{eq:pathConstraints} \\
 r^{\text{eq}} \! \left(x \! \left(0\right), \dots, x \! \left(h_{f}\right), p \right) &= 0 \;, & \label{eq:equalityConstraints} \\
 r^{\text{ineq}} \! \left(x \! \left(0 \right), \dots, x \! \left(h_{f}\right), p \right) &\geq 0 \;. & \label{eq:inequalityConstraints} 
\end{align}
\end{subequations}

The solution of this OCP yields the optimal state trajectories $x\!\left(t\right)$ (which consist of the optimal generalized positions $q\!\left(t\right)\in\mathbb{R}^{n_{\text{dof}}}$, velocities $\dot{q}\!\left(t\right)\in\mathbb{R}^{n_{\text{dof}}}$ and joint torques $\tau\!\left(t\right)\in\mathbb{R}^{n_{\text{act}}}$), the optimal control trajectories $u\!\left(t\right)=\dot{\tau} \!\left(t\right)\in\mathbb{R}^{n_{\text{act}}}$, parameters $p$ and phase durations $h_{0}, \dots, h_{9}$. 

As shown in and explained on Figure \ref{fig:overview}, the study is divided in two steps. The main difference is in the intent with which the OCP is formulated and, consequently, in the concrete formulation of the objective function \eqref{eq:objectiveFunction} which will be explained in Sections \ref{subsubsec:ocpLSQ} and \ref{subsubsec:ocpOCP}.

Equations \eqref{eq:ode} and \eqref{eq:switchingCondition} are placeholders for the full multiphase dynamics as described in Section \ref{subsubsec:modelingMathematical}. The path constraints \eqref{eq:pathConstraints} specify the upper and lower bounds on all variables included in the OCP. The upper and lower bounds for the transition phases are set to zero, as these phases have no length. For the dynamics reconstruction formulation, the limits were set generously to make accurate tracking of the reference motion possible. In the case of the motion synthesis formulation, bounds were deduced from the reconstructed motions. For a more realistic description, the limits on the joint torques are set for each node based on the current joint angle and angular velocity by inequality constraints \eqref{eq:inequalityConstraints}. The computation of the maximal extension and flexion torques is based on a muscle model after applying a fitting routine for the individual athletes as described in detail in \cite{Millard2013} (implemented as AddOn to RBDL \cite{Felis2017}). In addition, constraints specifying proper ground contact, touchdown and lift-off of the feet are formulated as equality and inequality constraints \eqref{eq:equalityConstraints} and \eqref{eq:inequalityConstraints} as well. The latter are further used to introduce landing conditions in the case of the motion synthesis problem formulation: We require that the athlete land on their heels and that constrain the distance between the feet (or the foot and the prosthesis, respectively), the distance between the hands as well as the distance between hands and feet.

For the solution of the optimal control problem, we employ the software package MUSCOD-II \cite{Bock1984,Leineweber2003} which uses a direct method for control discretization and a multiple shooting method for state parameterization. The result of these, a large non-linear programming problem, is treated by a specially tailored sequential quadratic programming method which applies condensing techniques for the handling of the underlying quadratic subproblems.

\subsubsection*{Objective Function for Dynamics Reconstruction}
\label{subsubsec:ocpLSQ}
The dynamics reconstruction problem aims at reconstructing the dynamic characteristics of the respective motion from purely kinematic reference data. The kinematic reference data are motion capture recordings of one athlete with BKA and one non-amputee athlete taken at the German Sport University Cologne. For a detailed description of the methods used during data capturing please refer to \cite{Willwacher2017b}. The c3d-files include marker trajectories and force plate data. The latter was not used in the reconstruction, but was only included in the analysis for comparison purposes. To transfer the marker positions onto our models which are described by their overall position and relative joint angles, we employ the tool Puppeteer \cite{Felis2015a} which is based on an inverse kinematics fitting routine. We refer to the so computed generalized positions for the subject-specific rigid multi-body models as reference motions.

For the dynamics reconstruction problem, we apply a least squares objective function of the form:

\begin{equation}
 \hspace*{1cm} \min_{x \left( \cdot \right), \, u \left( \cdot \right), \, p} \sum_{k = 0}^{m}
 {\frac{1}{2} \left( \left\| W \left( q_{k}^{MC} - q \left( h_{k} \right) \right) \right\|_{2}^{2} + \gamma_{u} \left
 \| u \! \left( h_{k} \right) \right\|_{2}^{2} \right)} \; .
\label{eq:objectiveLSQ}
\end{equation}

It has two contributing parts: The deviations between the generalized positions $q \! \left(h_{k}\right)$ of the model and the generalized reference positions $q_{k}^{MC}$ are evaluated at corresponding time points $h_{k}$ (the shooting nodes). To assess the correct time points, the reference motion is interpolated by splines, since the fixed sampling rate at which it is recorded might not coincide with the shooting node timings. As we compare very different orders of magnitudes (positions in \si{\m} and angles in \si{\radian}), a diagonal weight matrix $W\in\mathbb{R}^{n_{\text{dof}}}$ might be used for balancing. The second term which features the control variables is added to guarantee a unique solution and suppress excessive oscillations due to numerical issues. We choose the scaling factor $\gamma_{u}$ carefully to guarantee that the least squares term contributes mainly to the objective function and that the regularization term does not hide any actually existing (torque/force) phenomena.

\subsubsection*{Objective Function for Prediction of Motions}
\label{subsubsec:ocpOCP}
The objective function for motion generation OCPs can be written as a linear combination, 

\begin{equation}
 \Phi = \sum_{j=1}^{n_{M}} \gamma_j \phi_{M_j} \! \left( t_{f}, x\!\left(t_{f}\right), p \right) + \sum_{k=1}^{n_{L}} \gamma_k \sum_{i=1}^{N} \int_{t_{i-1}}^{t_{i}} \phi_{L_k} \! \left( t, x\!\left(t\right), u\!\left(t\right), p \right) \mathrm{d}t \; ,
 \label{eq:objectiveOCP}
\end{equation}

of $n_{M}$ Mayer type objective functions and $n_{L}$ Lagrange type objective functions. Each basic objective functions formalizes a single optimization criterion. We can distinguish criteria that are only evaluated at the end of a phase (Mayer type) and criteria that are evaluated at each node within a phase (Lagrange type). The coefficients $\gamma$ of the linear combination make the investigation of different objective functions possible as well as a weighting of the influence that the individual criteria have on the resulting movement. After an analysis of real long jumps and the results of the dynamics reconstruction, we have selected the following basic criteria for a more detailed examination:

\begin{enumerate}
 \item \emph{Maximization of jumping distance:} We compute the jumping distance $d_{\text{jump}}$ as distance between between the back end of the virtual board (located at $x=0$) and the landing position $P_{x}^{*} \! \left( h_{\text{landing}}\right)$. Hence, 
 \begin{subequations}
 \begin{equation}
 \phi_{M_0} \! \left( t_f, x \! \left( t_f \right), p \right) = - p_{\text{d}} = - P_{x}^{*} \! \left( h_{\text{landing}}\right) \; .
 \label{eq:jumpOptDistance} 
 \end{equation}
 
The asterisk denotes the heel contact point of the left foot in the case of the amputee model as the landing constraint requires a touchdown with this contact point. In the case of the non-amputee athlete, the asterisk stands for the heel contact point of the rear foot, i.e. the one with the shorter distance from the take-off board. 
 
 \item \emph{Minimization of Torque Derivatives Squared:} As in the dynamics reconstruction OCP, a minimization of the control variables is added as a regularization term to guarantee a unique solution:
 
 \begin{equation}
 \phi_{L_1} \! \left( t, x \! \left( t \right), u \! \left( t \right), p \right) = \left\| \dot{\tau} \! \left( t \right) \right\|^{2}_{2} = \left\| u \! \left( t \right) \right\|^{2}_{2} \; .
 \label{eq:jumpOptTorqueDerivs}
 \end{equation}
 
 \item \emph{Constant Head/Gaze Stabilization:} The head stabilizing Lagrange-type objective function minimizes uncontrolled wobbling of the head:
 
 \begin{equation}
 \phi_{L_2} \!  \left( t, x \! \left( t \right), u \! \left( t \right), p\right) = \left\| \theta_{\text{head,abs}} \right\|^{2}_{2} \; .
 \label{eq:jumpOptHeadStab}
 \end{equation}
 \end{subequations}

\end{enumerate}

\section*{Numerical Results}
\label{sec:results}
\begin{figure}[tbp]
 \centering
 \subfloat[Computed solutions for the athlete without BKA]{\label{fig:jump_sequence_NA} \includegraphics[width=0.95\textwidth]{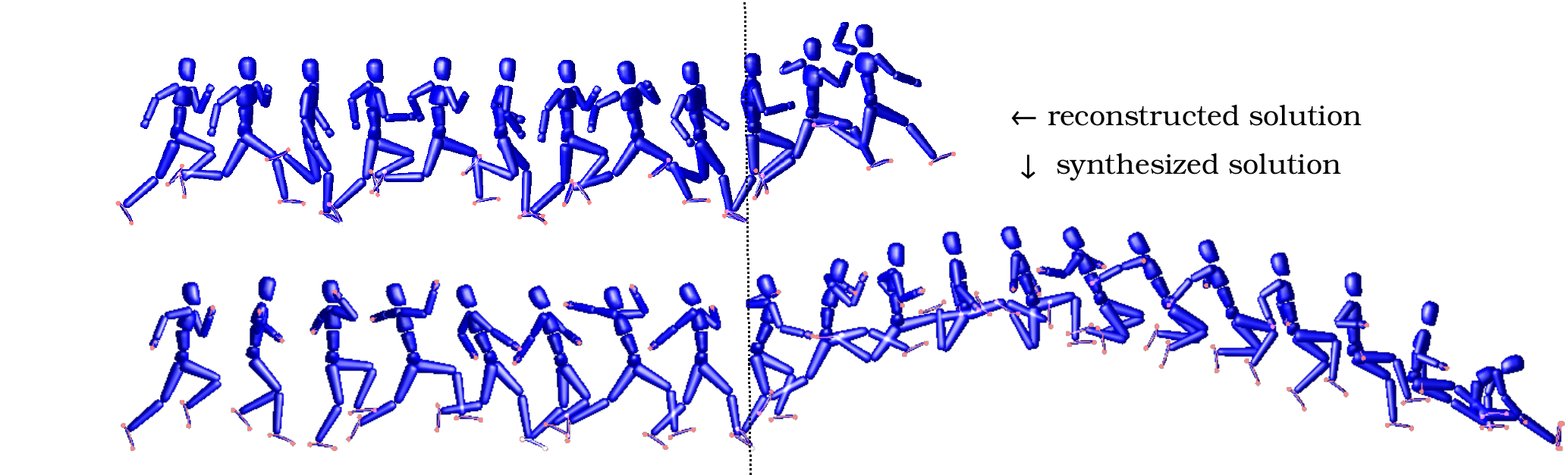}}\\
 \subfloat[Computed solutions for the athlete with BKA]{\label{fig:jump_sequence_A} \includegraphics[width=0.95\textwidth]{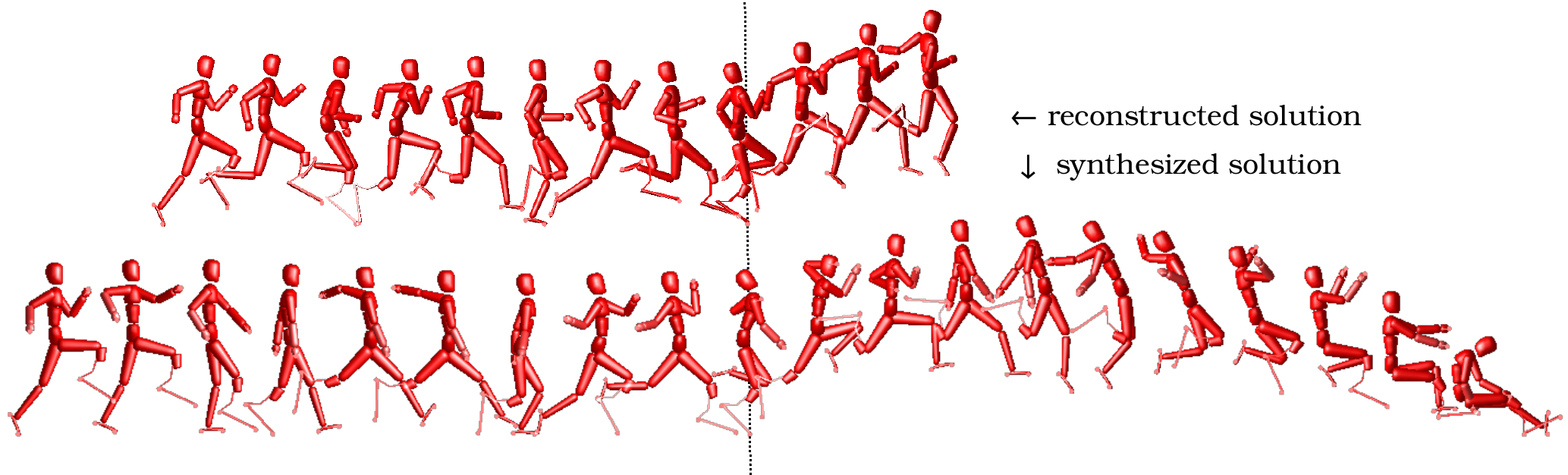}}
 \caption{Animated sequences of the optimal control problem generated long jump motions. The upper row shows the reconstructed and the lower row the synthesized motion. The dashed vertical line indicates the take-off position and motions are aligned to it. As immediately visible, the motion capture recordings for the reconstruction end during the jump.}
 \label{fig:jump_sequence}
\end{figure}

Figure \ref{fig:jump_sequence} shows the results of the dynamics reconstruction and the motion prediction of long jump motions for the athlete without BKA (Figure \ref{fig:jump_sequence_NA}) and the athlete with BKA (Figure \ref{fig:jump_sequence_A}). For the dynamics reconstruction, the movement ends during the flight where the motion capture data end.

To assess the quality of the dynamics reconstruction, we compute the root-mean-square error (RMSE) between the generalized positions of the OCP solutions $q_{k}^{OCP}$ and the reference data $q_{k}^{MC}$ by

\begin{equation}
 RMSE = \sqrt{\frac{1}{n_{q} m} \left( \sum_{k \in \mathcal{I}_{DOF}} \sum_{j=0}^{m} \left( q_{k}^{OCP} \! \left( t_j \right) - q_{k}^{MC} \! \left( t_j \right) \right)^{2} \right)} \; .
 \label{eq:rmse}
\end{equation}

The RMSE are computed separately for the translational and rotational degrees of freedom (DOF) by introducing a set of indices $\mathcal{I}_{DOF}$. The number of DOFs $n_q = |\mathcal{I}_{DOF}|$ and the number of multiple shooting nodes $m$ are used as normalization factors. Over the entire motions of both athletes in this study, the RMSE is smaller than \SI{2}{\cm} for the translational and smaller than \SI{0.036}{\radian} ($\approx \SI{2}{\degree}$) for the rotational DOF.

\begin{figure}[tbp]
\centering
 \subfloat[Athlete without BKA]{\label{fig:jumpReco_forces_NA}\includegraphics[width=0.95\textwidth]{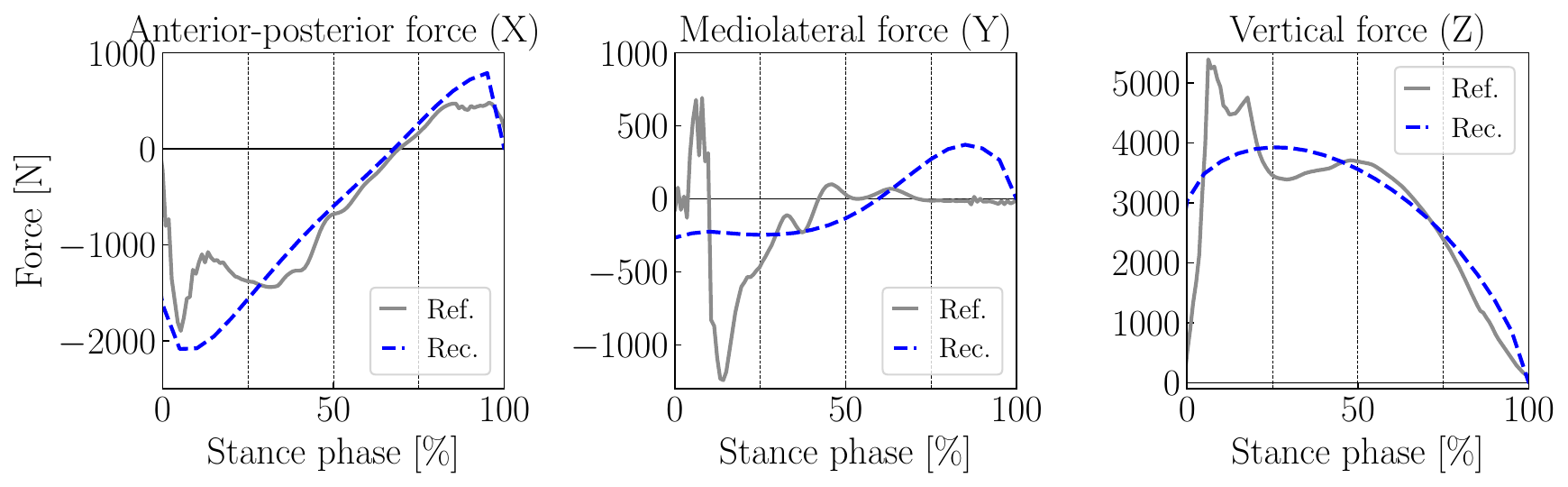}} \\
 \subfloat[Athlete with BKA]{\label{fig:jumpReco_forces_A}\includegraphics[width=0.95\textwidth]{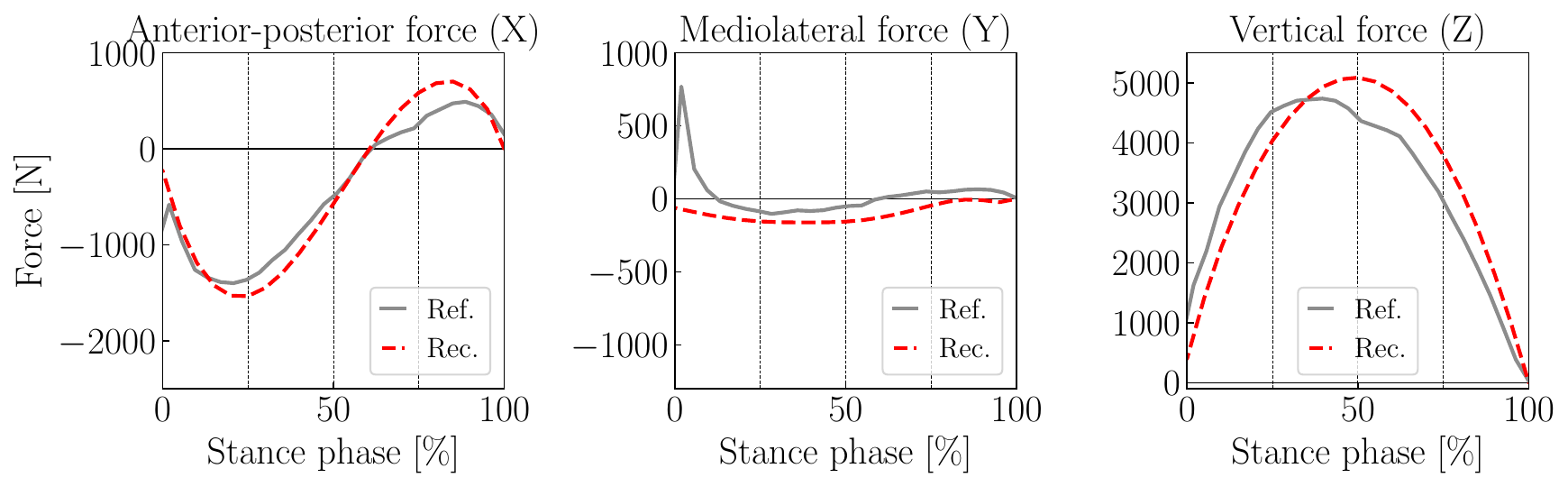}}
\caption{Anterior-posterior, mediolateral and vertical ground reaction forces during take-off for the reconstructedsilutions of the athlete without (a/blue) and with (b/red) BKA during the take-off. The grey lines show the measured raw force plate data and the colored lines the reconstructed forces. The time on the x-axis is given in percent of the stance time.}
\label{fig:jumpReco_forces}
\end{figure}

Figure \ref{fig:jumpReco_forces} shows the raw force plate data and the reconstructed ground reaction forces during take-off for the athlete without BKA (upper row) and the athlete with BKA (lower row). 
Altogether, the reconstructed and measured ground reaction forces fit well together, both in terms of course and magnitude.
For the anterior-posterior force (X), both models generate first a braking force component and then a propulsive force component. The transition between the two components occurs after the same percentage of the stance time. The reconstruction results in slightly larger peak values in both components.
For the mediolateral forces (Y), the reconstructed forces differ from the raw measurement data.
For the vertical forces (Z), the reconstructed solutions are of parabolic shape. In the measurement data for the non-amputee athlete, a sharp peak can be seen at the beginning of the contact, leading to a two-peak structure which does not show in the reconstructed solution.

\begin{figure}[tbp]
\centering
\includegraphics[width=0.95\textwidth]{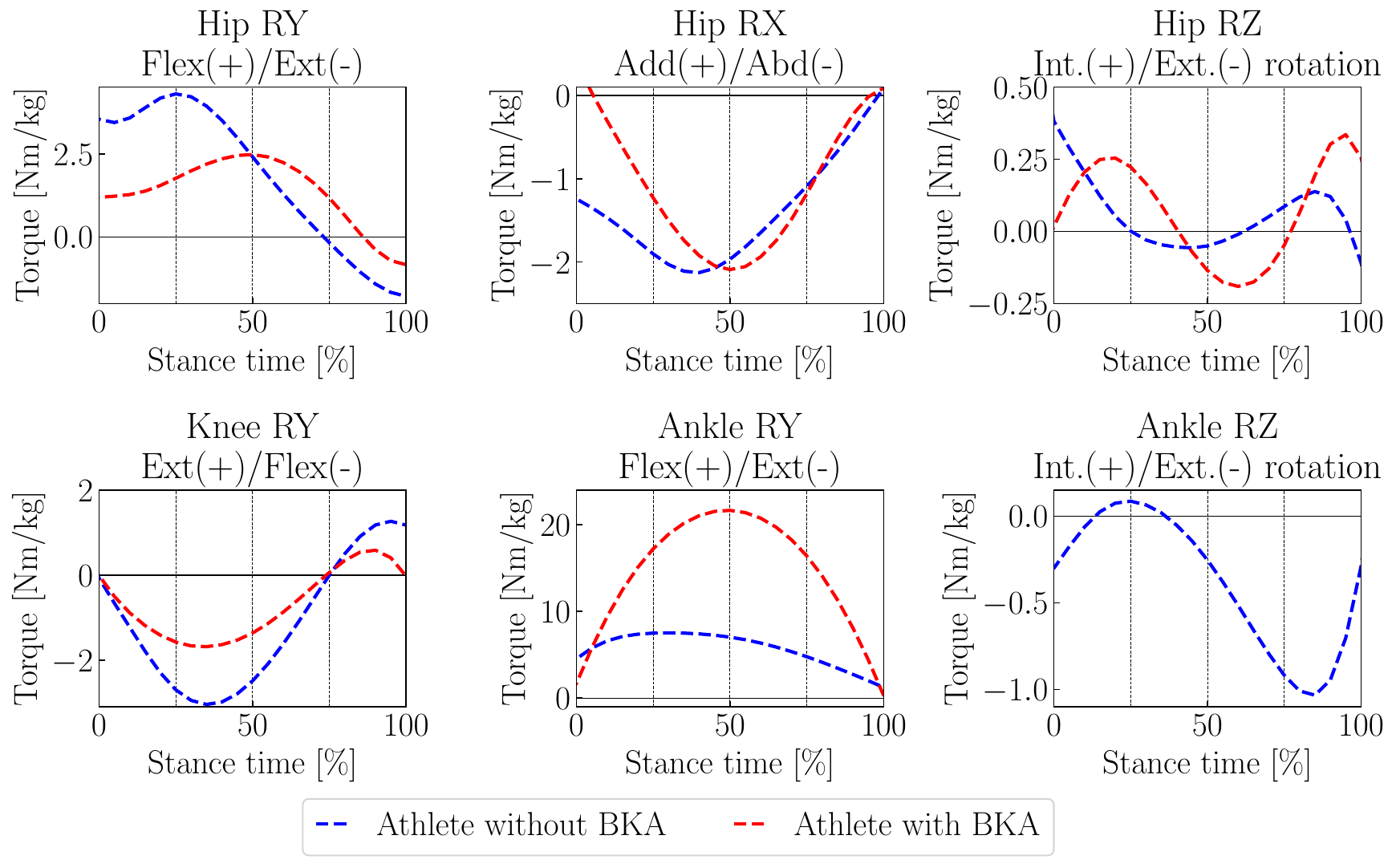}
\caption{Joint torques of the stance leg (right/affected leg) during the take-off for the reconstructed solutions of the athlete with (red) and the athlete without (blue) BKA. The time on the x-axis is given in percent of the stance time and the joint torques are normalized by body mass.}
\label{fig:jumpReco_tau_stance}
\end{figure}

Figure \ref{fig:jumpReco_tau_stance} shows the joint torques of the reconstructed solutions during take-off normalized by body mass. The joint torque generated passively in the prosthetic ankle is computed using Eq. \eqref{eq:springDamper}.

We compute an estimate of the jumping distance in analogy to the computation of Willwacher et al. \cite{Willwacher2017b} which is based on the assumption of a parabolic flight curve of the CoM taking into account the CoM positions and velocities at lift-off and landing. Then, we compute a jumping distance of \SI{7.58}{\m} for the athlete without and of \SI{8.29}{\m} for the athlete with BKA.
The take-off angle $\beta_{\text{take-off}}$ (computed as the arcustangens of the ratio of vertical and horizontal CoM velocity at lift-off) is \SI{16.38}{\degree} for the athlete without and \SI{20.25}{\degree} for the athlete with BKA.

For the motion synthesis solutions, we compute a jumping distance of \SI{9.86}{\m} with a take-off angle of \SI{20.09}{\degree} for the athlete without BKA and a jumping distance of \SI{9.22}{\m} with a take-off angle of \SI{16.85}{\degree} for the athlete with BKA. Hence, the jumping distances of the synthesized solutions are larger than the ones of the reconstructed solutions for both athletes (by \SI{2.28}{\m} for the athlete without BKA and by \SI{0.93}{\m} for the athlete with BKA). Regarding the take-off angles, we notice that the take-off angle of the synthesized solution is larger than the one of the reconstructed solution for the athlete without BKA (by \SI{3.71}{\degree}) and smaller for the athlete with BKA (by \SI{3.40}{\degree}).

\begin{figure}[tbp]
\centering
\includegraphics[width=0.95\textwidth]{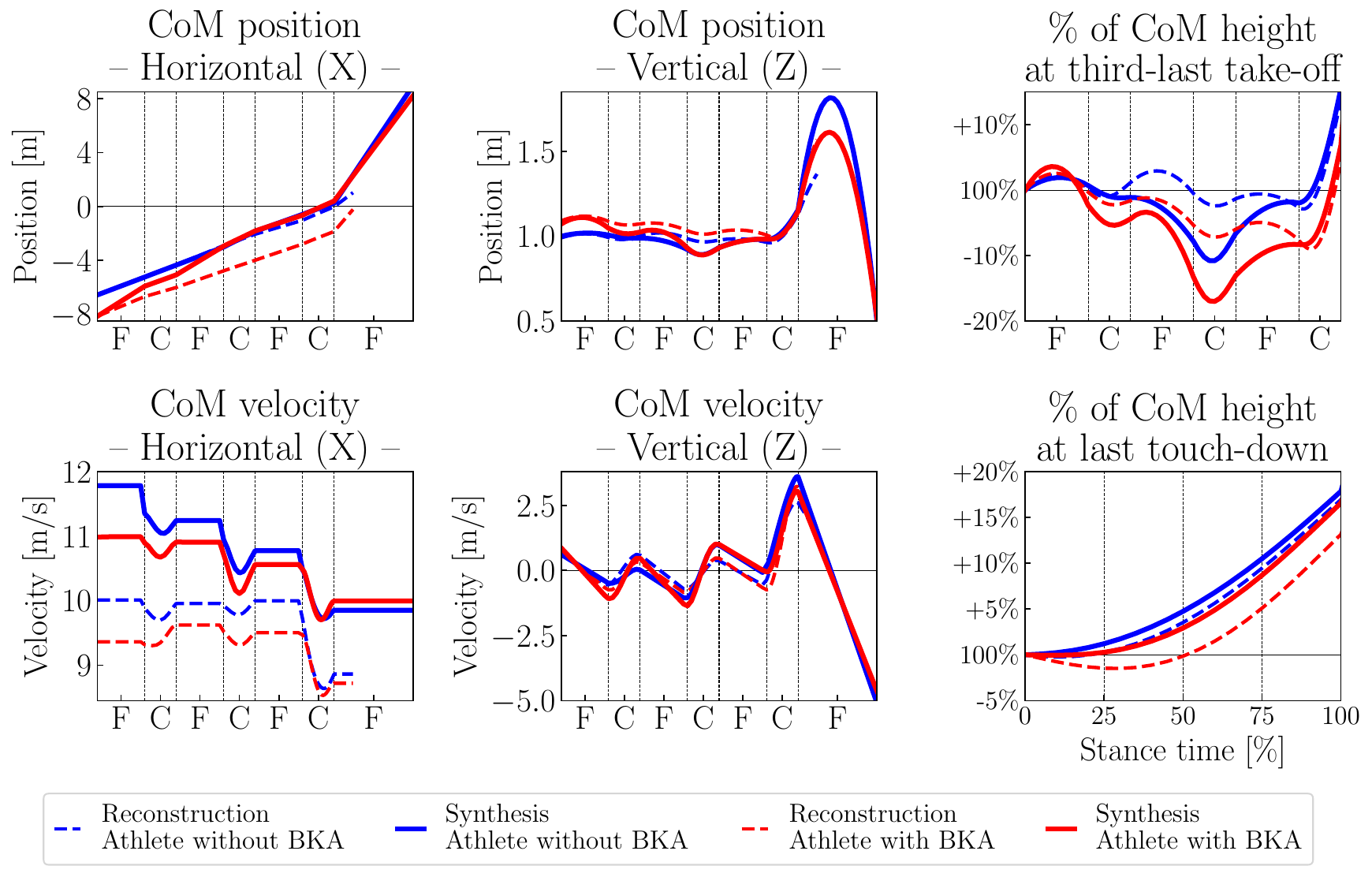}
\caption{The diagrams show the movement of the CoM in horizontal and vertical directions for the reconstructed (dashed lines) and synthesized (solid lines) long jump solutions for the athlete with (red) and without (blue) BKA. Phase durations are normalized individually per phase and the abbreviations \enquote*{F} and \enquote*{C} denote flight and contact phases (cf. Figure \ref{fig:phasesLongJump}).}
\label{fig:jump_com}
\end{figure}

Figure \ref{fig:jump_com} shows the motion of the CoM with six subdiagrams. The two left columns of the upper row show the horizontal and vertical CoM position and the two left columns of the lower row show the horizontal and vertical CoM velocities over the whole motion (with phase durations normalized per phase for better comparability). The right column is related to the development of the vertical CoM position with respect to its height at fourth-last take-off (upper row) and last touch-down (lower row). 
For both athletes, the synthesized horizontal CoM velocities are larger than the reconstructed ones. At the beginning of the investigated motion sequence, the difference between the velocities is \SI{1.77}{\m\per\s} for the athlete without BKA and \SI{1.63}{\m\per\s} for the athlete with BKA. At take-off, the difference between the velocities is \SI{0.99}{\m\per\s} for the athlete without BKA and \SI{1.27}{\m\per\s} for the athlete with BKA. In the synthesized solutions, the horizontal CoM velocities at take-off are \SI{9.85}{\m\per\s} for the athlete without BKA and \SI{9.99}{\m\per\s} for the athlete with BKA. While the horizontal CoM velocities of the reconstructed solutions remain nearly constant over the last steps before take-off for both athletes, we find a clear deceleration over the steps in the synthesized motions.    
In the vertical component of the CoM position, we observe a stronger lowering of the CoM in the synthesized solutions of both athletes compared to the reconstructed ones, especially during the second-last contact. Regarding the vertical CoM velocity, it is particularly noticeable that the gain in vertical velocity in the synthesized solutions is already significantly larger during the penultimate contact phase than in the reconstructed solution (by circa \SI{90}{\percent} for both athletes). Accordingly, in the synthesized soltions, both athletes start with a vertical velocity around zero in the take-off. For the athlete without BKA, this leads to a significantly larger vertical velocity at take-off (\SI{2.61}{\m\per\s} in the reconstructed and \SI{3.60}{\m\per\s} in the synthesized solution).

\begin{figure}[tbp]
\includegraphics[width=0.95\textwidth]{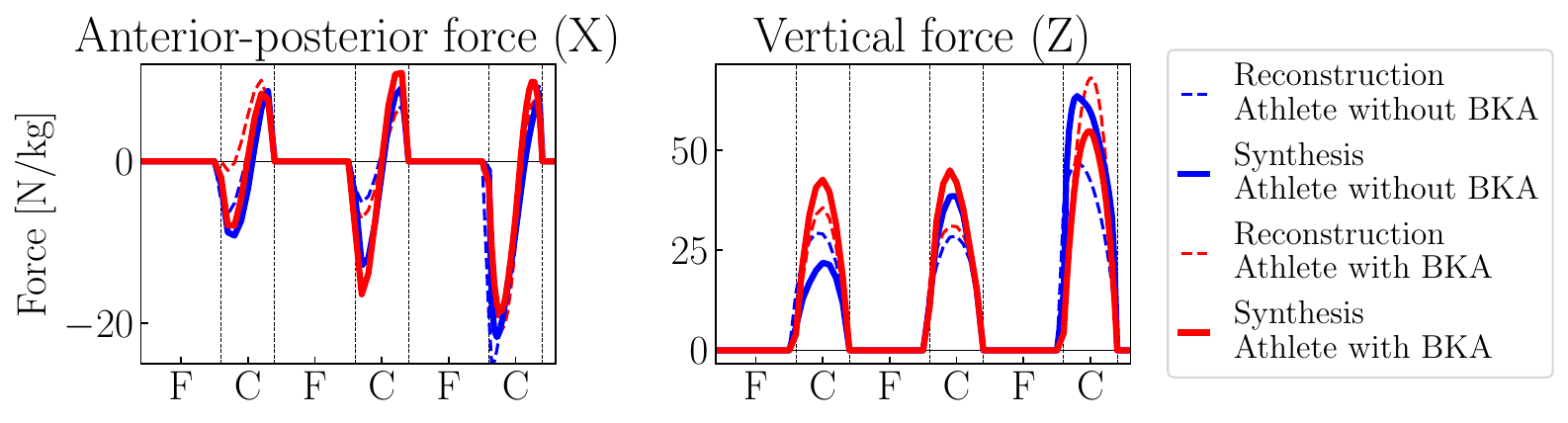}
\caption{Ground reaction forces in horizontal and vertical direction for the reconstructed (dashed lines) and synthesized (solid lines) long jump solutions for the athletes with (red) and without (blue) BKA. Each force is normalized by the body mass of the individual athlete. Phase durations are normalized individually per phase and the abbreviations \enquote*{F} and \enquote*{C} denote flight and contact phases (cf. Figure \ref{fig:phasesLongJump}).}
\label{fig:jump_forces}
\end{figure}

Figure \ref{fig:jump_forces} gives the horizontal and vertical ground reaction forces for the last two steps of the long jump approach and the take-off. For a better comparability between the two athletes, we normalize the forces by body mass.
Differences are particularly evident in the vertical component of the ground reaction forces -- both when comparing the reconstruction and synthesis and when comparing the two athletes. For the athlete without BKA, the synthesized force is smaller than the reconstructed force during the third-to-last contact and larger for the last two contacts. For the amputee athlete, the synthesized force during the last contact (take-off) is smaller than in the reconstructed solution and larger during the previous two contacts. In the reconstruction, the normalized ground reaction force of the athlete with BKA is larger than that of the athlete without BKA in each comparison. In the synthesis, the force of the athlete with BKA is smaller than that of the athlete without BKA during the last contact (take-off) and larger during the two previous contacts. The synthesized solution also misses the two-peak structure with an active and a passive peak in the solution for the athlete without BKA.

\begin{figure}[tbp]
\centering
\includegraphics[width=0.95\textwidth]{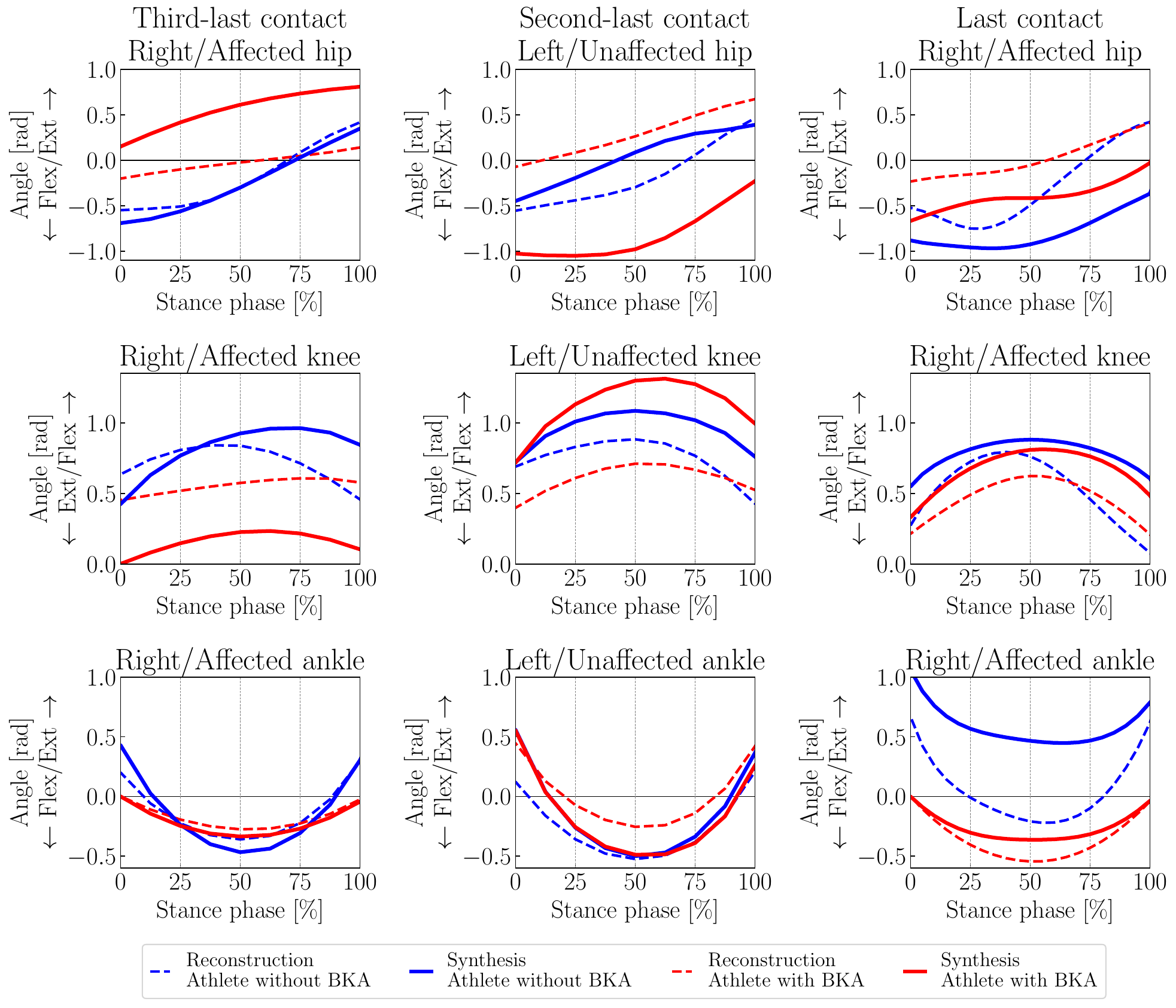}
\caption{Joint angles of the respective contact leg for the three contact phases of the reconstructed (dashed lines) and synthesized (solid lines) long jump solutions for the athletes with (red) and without (blue) BKA. Phase durations are normalized to stance time.}
\label{fig:jump_q}
\end{figure}

\begin{figure}[tbp]
\centering
\includegraphics[width=0.95\textwidth]{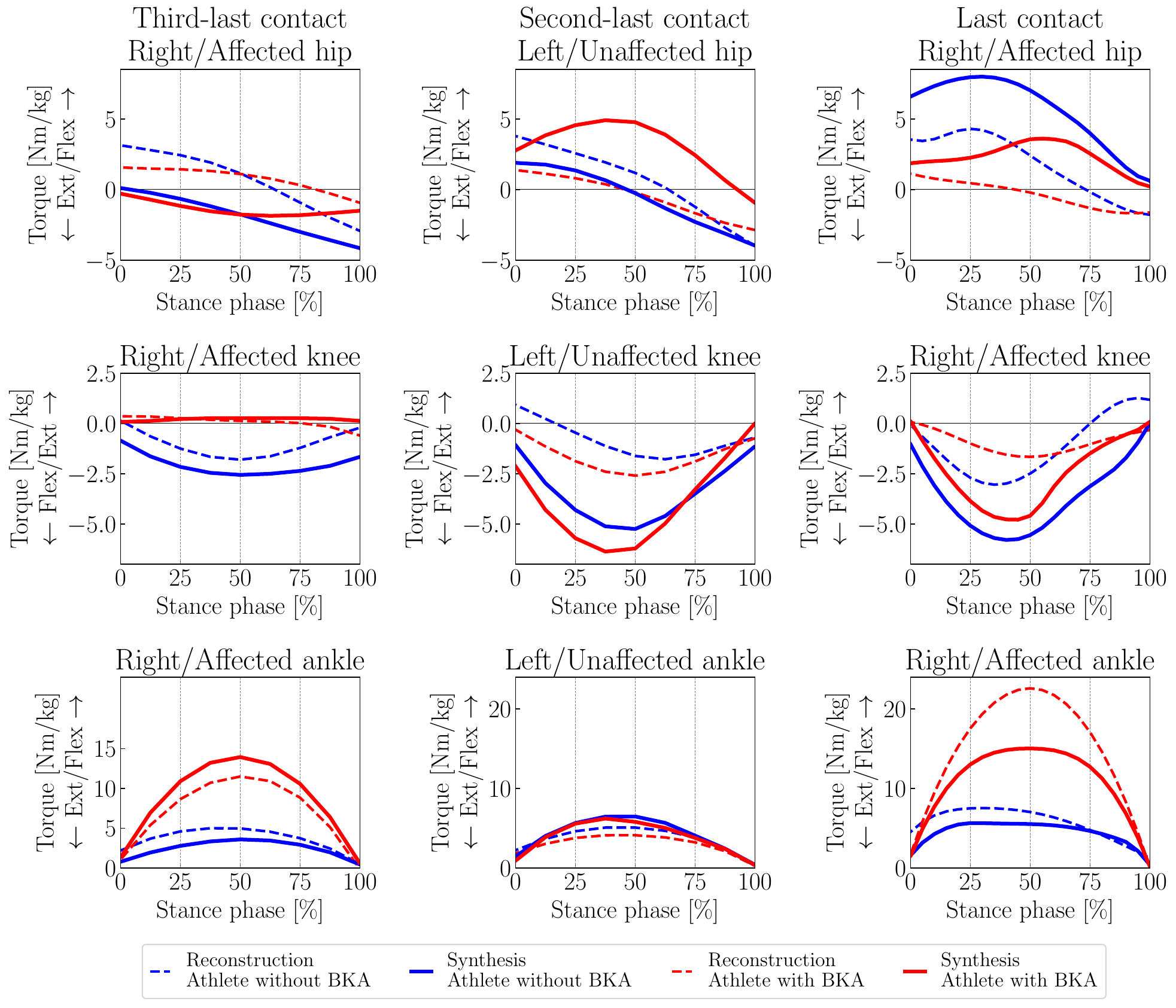}
\caption{Joint torques of the respective contact leg for the three contact phases of the reconstructed (dashed lines) and synthesized (solid lines) long jump solutions for the athletes with (red) and without (blue) BKA. Phase durations are normalized to stance time. Each torque is normalized by the body mass of the individual athlete. The passively generated joint torque of the prosthetic ankle is computed via Eq. \eqref{eq:springDamper}.}
\label{fig:jump_tau}
\end{figure}

Figures \ref{fig:jump_q} and \ref{fig:jump_tau} show the joint angles and torques of the respective contact leg (ankle, knee, hip) for the three contact phases (last two steps of approach and take-off). In both figures, the upper row shows the curves for the hip joint, the middle row the ones for the knee joint and the lower row the ones for the ankle joint. To make the differences between the individual contact phases easier to recognize, the y-axes are scaled consistently for each row. Like the forces, the joint torques are also normalized to the body mass of the individual athletes. The joint torque in the prosthetic ankle is computed based on Eq. \eqref{eq:springDamper} with the spring constant $k=\SI{3150}{\N\m\per\radian}$ reconstructed as a parameter of the dynamics reconstruction OCP. 
For both athletes, we observe significant differences in the joint angles and torques during the first contact phase of the motion (e.g., knee joint angle for athlete with BKA or hip joint torque for athlete without BKA). For the athlete without BKA, there is less knee flexion in the synthesized solution than in the reconstructed one despite larger flexion torques. The ankle joint torques of the synthesized solution are smaller during contact with the right leg and larger during contact with the left leg compared to the reconstructed solution. For the athlete with BKA, the knee flexion is similar between reconstructed and synthesized solutions. In contrast, the ankle flexion is larger in the synthesized solutions during the approach contact phases and smaller during the take-off with correspondingly larger (approach) or smaller (take-off) joint torques in the ankle.   

\section*{Discussions}
\label{sec:discussion}
The first purpose of our research was to evaluate the correctness of the proposed long jump model by computing the dynamics from purely kinematic motion capture data and comparing the results to force plate measurements and literature data. The small RMSE show that the reconstructed kinematics fits the motion capture data well. The RMSE can mainly be explained by the fact that the OCP formulation requires physically correct ground contact, which may not necessarily be present in the reference motion. However, we want to emphasize that these errors represent the deviations resulting from the optimal control step. Further deviations between the OCP solution and the actual movement can occur due to measurement inaccuracies and errors during the transfer of the marker positions to the model (see Section \ref{subsubsec:ocpLSQ}). 

In the overall picture, the ground reaction forces and joint torques also agree well with the force plate data and literature values for joint torques \cite{Funken2019c}. In the ground reaction forces, one striking difference should be discussed here: the absence of the second peak (passive peak) in the vertical ground reaction force of the athlete without BKA. We assume that the rather simple foot-contact model with one fixed contact point at the ball of the foot is the main cause for this difference. Furthermore, the rigidity of the model with the torque actuators as well as the instantaneous touchdown may not be able to reproduce fast time scale actions such as the sharp peak in the vertical ground reaction force (see e.g. \cite{Seyfarth1999}). Regarding the comparison of the joint torques to literature values \cite{Funken2019c}, we notice that the reconstructed joint torques are slightly larger than the ones reported in literature. They are also smoother, i.e., more specific motions like the double peak at the beginning of the hip flexion (RY) moment or the peaks at the beginning in the hip abduction (RX) and hip external rotation (RZ) moments are not reconstructed. This is probably due to the fact that the problem formulation computes an optimal solution for the entire motion, thus smoothing out stronger \enquote{oscillations}

In total, the reconstructed solutions are in good accordance with measurements of long jump motions with and without RSP and hence allow using the proposed long jump model for further analysis, motion synthesis and comparison.

As shown in Figure \ref{fig:overview}, the results allow two different comparisons: one between the reconstructed and synthesized solution and one between the movements of the athlete with BKA and the athlete without BKA. Let's start with the first comparison, the one between the reconstructed and the synthesized solution, with the aim to work out to what extent the chosen objective function can generate realistic long jump movements and where the investigated athletes could still optimize their movements for a better performance.

Looking at the animated sequences in Figure \ref{fig:jump_sequence} shows that the synthesized solutions look quite realistic, even if differences to the reconstruction are already visible here. These become clearer when comparing the curves in Figures \labelcref{fig:jump_com,fig:jump_forces,fig:jump_q,fig:jump_tau}.

For both athletes, the synthesized solution calculated a significantly larger jumping distance than that of the reconstruction (and thus also the measurement of a real long jump). One reason for this is certainly intrinsic to the problem formulation as an optimal control problem: This formulation entails that the motion is considered as a whole with the aim of adjusting every variable exactly in such a way that the objective function becomes optimal. Hence, all variables are known at each time point and can be chosen with the knowledge of all previous and subsequent time points, which is not the case for a real athlete in a motion capture recording. Furthermore, however, the greater jumping distance also shows potential for improvement for the two individual athletes, and an examination of the movement parameters can show where changes can be made.

Since the jumping distance of both athletes increased in the optimal solution, but not the take-off angles (only for the athlete without BKA), there seems to be no direct (linear) relationship between the two quantities (even if they are certainly linked).

The changes in the CoM motion of the synthesized solutions compared to the reconstruction have a decisive influence on the larger jumping distance. Both athletes have a larger approach velocity in the synthesized solution compared to the measurements and manage to loose less forward CoM velocity during take-off. At this point, the question remains open whether the real athletes could actually implement the higher approach velocities calculated by the model with their muscular prerequisites. If not, this would be a possible starting point for specific training. 
A lowering of the CoM during the last steps of the approach is described in the literature \cite{Hay1986b} as a technique to bring the body into an appropriate position for the take-off contact phase in order to gain more vertical CoM velocity. It seems that the optimal solution has found a way to exploit this technique more than the real athletes. The differences in vertical CoM velocity between reconstruction and synthesis are also related to this greater lowering. It seems that the greater lowering allows to start with a vanishing vertical velocity in the take-off (instead of a negative vertical velocity), which enables a greater gain in vertical CoM velocity as well as a high vertical velocity in the take-off. 

The absence of the second peak in the vertical ground reaction force of the athlete without BKA during take-off also in the synthesized solution can be explained by similar arguments as already used for the reconstruction: a rather simple foot model with one contact point, numerical reasons (too few grid points) and the fact that tendon elasticity is not included in our formulation.

For the athlete without BKA, it is further noticeable that the peak value of the vertical force of the synthesized solution is smaller than in the reconstructed solution during the first contact phase, while it is larger than in the reconstructed solution in the other two contact phases. This is noticeable because it would be expected that the vertical ground reaction forces in an optimized solution focusing on the maximization of a criterion would always be greater than or comparable to those in reality. Together with the differences between the reconstructed and synthesized solution in the joint angles and joint torques of both athletes, which are especially large in this first contact phase, the question arises as to the cause of this unexpected behavior. First, it would be possible that more degrees of freedom in the knee and the prosthesis are necessary for a meaningful calculation of the reconstructed long jump motion and that the model assumptions made have a negative effect at this point. Since the synthesized solution was calculated with a model additionally reduced in the number of degrees of freedom, this may propagate. On the other hand, it would be possible that the effect arises from the fact that the affected contact phase is located at the beginning of the investigated motion sequence and would disappear if further previous steps were considered. For the other two contact phases, the behaviour of the solutions is more as expected: The curves for the joint angles and joint torques of the athlete without BKA show a similar course, but differ in magnitude and absolute values with the synthesized torques often being larger regarding absolute values.

For the athlete with BKA, it is noticeable that the vertical ground reaction force during take-off is significantly smaller in the synthesized solution compared to the reconstructed solution; the peak force is about \SI{24}{\percent} smaller here.  In contrast, the vertical force of the synthesized solution during the contact of the prosthetic device in the approach is larger than in the reconstructed solution which is in accordance with optimization-based studies on sprinting motions with RSP where the optimized solutions make greater use of the spring-like properties of the RSP to produce large vertical ground reaction forces \cite{Emonds2019b}. This observation is reflected in the ankle angles and torques, as well. During the contact phases of the approach, there is a similar or larger flexion  and a larger corresponding torque in the synthesized solution compared to the reconstructed solution. During take-off, it is exactly the opposite: less flexion and a smaller joint torque in the synthesized solution. Thus, the RSP is used completely differently in reality (described by the reconstruction of the measurements) and in synthesis: While the athlete with BKA uses the RSP more during (approach) sprinting steps in the optimal solution, the energy-storing properties are used less during take-off. This raises the question of why the optimized solution shows a significantly different take-off behavior here for achieving a greater jumping distance. In this context, it is further interesting that the intentional stiffening of the knee described in the literature is not present in the optimized solution. In fact, the solution calculates a significant flexion of the knee and, correspondingly, a significantly larger flexion torque than in the reconstructed solution. 

In summary, there are discrepancies between the reconstructed and synthesized solutions, in some places significant ones. However, basic biomechanical principles of running and jumping are present in both athletes and both calculation methods, as differences are primarily due to different magnitudes of variables or detail deviations. Hence, we think that a comparison of the synthesized motions of the athlete with BKA and the athlete without BKA is valid and can provide new interesting insights. Before moving on to this comparison, we want to explain possible reasons for the deviations between reconstructed and synthesized solutions:

\begin{enumerate}
    \item It is expected that an optimized movement deviates in some aspects from a recorded motion. The synthesis solution is optimal with respect to the chosen criteria and movement model, whereas a real athlete must execute his movements intuitively and based on training experience in a fraction of a second and react to disturbances. Moreover, the synthesis computes the solution as a whole with knowledge of all variables at all time points and according to the possibility of choosing optimal positions at all time points.
    \item It is possible that the synthesis solution is optimal with respect to the optimization criteria and physically feasible, but not intuitive or too complex for execution by a real athlete. 
    \item It is possible that the selected optimization criteria or the problem formulation need to be improved. Optimization criteria on the take-off could be, for example, aiming at a certain take-off angle or maximizing CoM velocities at take-off. Here, an intensive study of different combinations of criteria could provide more insights. 
\end{enumerate}

Let's conclude the discussion with a comparison between the synthesized solutions of the athlete with BKA and the athlete without BKA. In direct comparison, the estimated jumping distance of the athlete without BKA is \SI{64}{\cm} longer than the one of the athlete with BKA. Hence, the real athlete without BKA might have even more potential to improve performance than the athlete with BKA. Since we compare only two distinct athletes, we cannot generalize the observation to jumping with and without RSP. Instead, we investigate which differences in the synthesized long jump motions induce the jumping distance differences. The most crucial differences are found in CoM motion and ground reaction forces during take-off.

In the literature (e.g. \cite{Hay1986b}), a high approach velocity and the lowering of the CoM during the last steps before take-off are described as important criteria for the long jump approach. Based on this, the athlete with BKA starts the take-off phase with better preconditions than the athlete without BKA due to a stronger lowering of the CoM height and a high horizontal CoM velocity. During the take-off this picture changes, because the athlete without BKA uses the take-off more efficiently: The athlete without BKA generates a larger vertical ground reaction force (with a \SI{16}{\percent} larger peak value than the athlete with BKA) in a shorter contact time (by \SI{20}{\percent}) resulting in a larger vertical change of momentum (by \SI{3.4}{\percent}) and a larger gain in vertical CoM velocity (by \SI{18.7}{\percent}). Due to the stronger lowering in CoM height before and a comparable large gain in CoM height during take-off, the athlete with BKA takes off with a lower CoM height than the athlete without BKA. It seems that a too strong lowering of CoM height is not advantageous for long jump performance. The observation that the athlete without BKA has a more efficient take-off technique in the synthesized solution than the athlete with BKA is particularly interesting because it is in contrast to the conclusions from Willwacher and colleagues \cite{Willwacher2017b} that athletes with BKA jump off more efficiently.
%
\section*{Conclusions}
\label{sec:conclusions}
To conclude this work, we summarize the main findings and place them in the larger context regarding long jump with and without RSP.
First, we can answer the research question with: Yes, the proposed long jump model is capable of generating physically correct and realistic long jump motions.  The comparison of the variables of the reconstructed solutions (which are an image of the measured motion) and synthesized solutions show similarities in the overall picture. However, there are still discrepancies in the details between them. We discussed several explanations for the remaining differences and it is likely that all explanation approaches have an influence.  

For both the athlete with and the athlete without BKA, the jumping distances of the synthesized solutions are larger than the respective ones of the reconstructed solutions. Thus, the optimized solutions include ways for both athletes to individually improve their performance. Our computations suggest that the athlete without BKA might have the capability to make the take-off more efficient and that the athlete with BKA might be able to approach the board with a larger horizontal velocity. Although the essential parts of a long jump occur mainly in the saggital plane, full three-dimensional motion synthesis might add useful insights. In particular, the arms and upper body are probably used in a balancing manner to stabilize the movement.

In the reconstructed solutions, the jumping distance of the athlete with BKA is larger than the one of the athlete without BKA. In contrast, it is the opposite for the synthesized solutions: the jumping distance of the athlete with BKA is shorter compared to the athlete without BKA. Since the comparison is made for two distinct athletes, it is possible that those differences are based on the different body and muscle parameters of the two athletes. Hence, a comparison of the athlete with BKA to an identical model without BKA is necessary to investigate whether the results show opportunities for performance improvement for the individual athletes or additionally suggest that the RSP has a hindering effect on the athlete with BKA achieving a larger jumping distance compared to the athlete without BKA. While not possible in measurements, such a modeling and comparison is indeed possible in the proposed computer model of long jump. To improve the results even more, accurate measurements of the muscular parameters of the athlete (e.g. using Biodex measurements) are necessary. Furthermore, a computer model comparing the athlete with BKA to a non-amputee version of the identical model allows for examining the influence of changes in individual parameters (e.g., prosthesis stiffness, muscle parameters) on the movement with RSP. These information can be used for training advice and adjustments.

\bibliography{references}

@Article{Alexander1990,
  author  = {Alexander, R.~M. and Bone, Q.},
  title   = {Optimum take-off techniques for high and long jumps},
  journal = {Philosophical Transactions of the Royal Society of London. Series B, Biological Sciences},
  year    = {1990},
  volume  = {329},
  number  = {1252},
  pages   = {3--10},
  doi     = {10.1098/rstb.1990.0144},
}

@Article{Funken2019c,
  author  = {Funken, J. and Willwacher, S. and Heinrich, K. and M{\"u}ller, R. and Hobara, H. and Grabowski, A.~M. and Potthast, W.},
  title   = {{Three-Dimensional Takeoff Step Kinetics of Long Jumpers with and without a Transtibial Amputation}},
  journal = {Medicine and Science in Sports and Exercise},
  year    = {2019},
  volume  = {51},
  number  = {4},
  pages   = {716--725},
  doi     = {10.1249/MSS.0000000000001853},
}

@Article{Hay1986b,
  author  = {Hay, J.~G.},
  title   = {{The Biomechanics of the Long Jump}},
  journal = {Exercise and Sport Sciences Reviews},
  year    = {1986},
  volume  = {14},
  pages   = {401--416},
}

@Article{Hay1993,
  author  = {Hay, J.~G.},
  title   = {{Citius, altius, longius (faster, higher, longer): The biomechanics of jumping for distance}},
  journal = {Journal of Biomechanics},
  year    = {1993},
  volume  = {26},
  number  = {Suppl 1},
  pages   = {7--21},
}

@Article{Lees1993,
  author    = {Lees, A. and Fowler, N. and Derby, D.},
  title     = {A biomechanical analysis of the last stride, touch‐down and take‐off characteristics of the women's long jump},
  journal   = {Journal of Sports Sciences},
  year      = {1993},
  volume    = {11},
  number    = {4},
  pages     = {303-314},
  doi       = {10.1080/02640419308730000},
  publisher = {Routledge},
}

@Article{Lees1994,
  author    = {Lees,A. and Graham-Smith, P. and Fowler, N.},
  title     = {{A Biomechanical Analysis of the Last Stride, Touchdown, and Takeoff Characteristics of the Men's Long Jump}},
  journal   = {Journal of Applied Biomechanics},
  year      = {1994},
  volume    = {10},
  number    = {1},
  pages     = {61 - 78},
  address   = {Champaign IL, USA},
  publisher = {Human Kinetics, Inc.},
}

@InBook{Nixdorf1990,
  chapter   = {Biomechanical analysis of the long jump},
  title     = {Scientific Research Project at the Games of the XXIVth Olympiad — Seoul 1988 Final Report},
  publisher = {International Athletic Foundation},
  year      = {1990},
  author    = {Nixdorf, E. and Brüggemann, G.-P.},
  editor    = {Brüggemann, G.-P. and Glad, B.},
  address   = {Monaco},
}

@InBook{Linthorne2008,
  chapter   = {Biomechanics of the long jump},
  pages     = {340--353},
  title     = {Routledge Handbook of Biomechanics and Human Movement Science},
  publisher = {Taylor \& Francis},
  year      = {2008},
  author    = {Linthorne, N.~P.},
  editor    = {Youlian Hong and Roger Bartlett},
}

@Article{Nolan2012,
  author  = {Nolan, L. and Patritti, B.~L. and Simpson, K.~J.},
  title   = {Effect of take-off from prosthetic versus intact limb on transtibial amputee long jump technique},
  journal = {Prosthetics and Orthotics International},
  year    = {2012},
  volume  = {36},
  number  = {3},
  pages   = {297--305},
  doi     = {10.1177/0309364612448877},
}

@Article{Seyfarth1999,
  author  = {Seyfarth, A. and Friedrichs, A. and Wank, V. and Blickhan, R.},
  title   = {Dynamics of the long jump},
  journal = {Journal of Biomechanics},
  year    = {1999},
  volume  = {32},
  number  = {12},
  pages   = {1259--1267},
  issn    = {0021-9290},
  doi     = {10.1016/S0021-9290(99)00137-2},
}

@Article{Willwacher2017b,
  Title                    = {Elite long jumpers with below the knee prostheses approach the board slower, but take-off more effectively than non-amputee athletes},
  Author                   = {Willwacher, S. and Funken, J. and Heinrich, K. and M\"uller, R. and Hobara, H. and Grabowski, A.~M. and Br\"uggemann, G.-P. and Potthast, W.},
  Journal                  = {Scientific Reports},
  Year                     = {2017},
  Volume                   = {7},
  Number                   = {1},
  Pages                    = {16058},
  Doi                      = {10.1038/s41598-017-16383-5},
}

@Article{Funken2019a,
  author  = {Funken, J. and Willwacher, S. and Heinrich, K. and M\"uller, R. and Hobara, H. and Grabowski, A.~M. and Potthast, W.},
  title   = {Long jumpers with and without a transtibial amputation have different three-dimensional centre of mass and joint take-off step kinematics},
  journal = {Royal Society Open Science},
  year    = {2019},
  volume  = {6},
  number  = {4},
  pages   = {190107},
  doi     = {10.1098/rsos.190107},
}

@Article{McGowan2012,
  Title                    = {Leg stiffness of sprinters using running-specific prostheses},
  Author                   = {McGowan, C.~P. and Grabowski, A.~M. and McDermott, W.~J. and Herr, H.~M. and Kram, R.},
  Journal                  = {Journal of The Royal Society Interface},
  Year                     = {2012},
  Number                   = {73},
  Pages                    = {1975--1982},
  Volume                   = {9},
  Doi                      = {10.1098/rsif.2011.0877}
}

@Article{Weyand2010,
  author  = {Weyand, P.~G. and Bundle, M.~W.},
  title   = {{Point: Artificial limbs do make artificially fast running speeds possible}},
  journal = {Journal of Applied Physiology: Respiratory, Environmental and Exercise Physiology},
  year    = {2010},
  volume  = {108},
  number  = {4},
  pages   = {1011--1012},
  doi     = {10.1152/japplphysiol.01238.2009},
}

@article{Emonds2019a, 
  title                    = {Comparison of Sprinting With and Without Running-Specific Prostheses Using Optimal Control Techniques}, 
  author                   = {Emonds, A.~L. and Funken, J. and Potthast, W. and Mombaur, K.}, 
  volume                   = {37}, 
  DOI                      = {10.1017/S0263574719000936}, 
  number                   = {12}, 
  journal                  = {Robotica}, 
  publisher                = {Cambridge University Press}, 
  year                     = {2019}, 
  pages                    = {2176--2194}
}

@article{Emonds2019b,
author = {Emonds, Anna Lena and Mombaur, Katja},
title  = {{Optimality Studies of Human Sprinting Motions with and Without Running-Specific Prostheses}},
journal = {International Journal of Humanoid Robotics},
volume = {16},
number = {03},
pages = {1940003},
year = {2019},
doi = {10.1142/S0219843619400036},
}

@Article{DeLeva1996,
  author  = {de Leva, P.},
  title   = {{Adjustments to Zatsiorsky-Seluyanov's segment inertia parameters}},
  journal = {Journal of Biomechanics},
  year    = {1996},
  volume  = {29},
  number  = {9},
  pages   = {1223--1230},
  doi     = {0021-9290(95)00178-6},
}

@Article{Schultz2010,
  Title                    = {{Modeling and Optimal Control of Human-Like Running}},
  Author                   = {Schultz, G. and Mombaur, K.},
  Journal                  = {IEEE/ASME Transactions on Mechatronics},
  Year                     = {2010},
  Number                   = {5},
  Pages                    = {783--792},
  Volume                   = {15},

  Doi                      = {10.1109/TMECH.2009.2035112}
}

@InBook{Mombaur2014,
  Title                    = {Modeling, Simulation and Optimization of Complex Processes - HPSC 2012},
  Author                   = {Mombaur, K.},
  Chapter                  = {{A Mathematical Study of Sprinting on Artificial Legs}},
  Editor                   = {Bock, H.~G. and Hoang, X. and Rannacher, R. and Schlöder, J.},
  Pages                    = {157--168},
  Publisher                = {Springer, Cham},
  Year                     = {2014},

  Doi                      = {10.1007/978-3-319-09063-4\_13}
}

@Article{Felis2017,
  Title                    = {{RBDL}: an efficient rigid-body dynamics library using recursive algorithms},
  Author                   = {Felis, M.~L.},
  Journal                  = {Autonomous Robots},
  Year                     = {2017},
  Number                   = {2},
  Pages                    = {495--511},
  Volume                   = {41},

  Doi                      = {10.1007/s10514-016-9574-0}
}

@Article{Millard2013,
  author  = {Millard, M. and Uchida, T. and Seth, A. and Delp, S.~L.},
  title   = {{Flexing Computational Muscle: Modeling and Simulation of Musculotendon Dynamics}},
  journal = {Journal of Biomechanical Engineering},
  year    = {2013},
  volume  = {135},
  number  = {2},
  doi     = {10.1115/1.4023390},
}

@Article{Bock1984,
  Title                    = {A multiple shooting algorithm for direct solution of optimal control problems},
  Author                   = {Bock, H.~G. and Plitt, K.~J.},
  Journal                  = {IFAC Proceedings Volumes},
  Year                     = {1984},
  Number                   = {2},
  Pages                    = {1603--1608},
  Volume                   = {17},

  Doi                      = {10.1016/S1474-6670(17)61205-9}
}

@Article{Leineweber2003,
  author  = {Leineweber, D.~B. and Bauer, I. and Bock, H.~G. and Schl{\"o}der, J.~P.},
  title   = {{An efficient multiple shooting based reduced SQP strategy for large-scale dynamic process optimization (Parts 1 and 2)}},
  journal = {Computers and Chemical Engineering},
  year    = {2003},
  volume  = {27},
  number  = {2},
  pages   = {157--174},
  doi     = {10.1016/S0098-1354(02)00158-8,
10.1016/S0098-1354(02)00195-3},
}

@InProceedings{Felis2015a,
  Title                    = {An optimal control approach to reconstruct human gait dynamics from kinematic data},
  Author                   = {Felis, M.~L. and Mombaur, K. and Berthoz, A.},
  Booktitle                = {2015 IEEE-RAS 15th International Conference on Humanoid Robotics (Humanoids)},
  Year                     = {2015},
  Pages                    = {1044--1051},

  Doi                      = {10.1109/HUMANOIDS.2015.7363490}
}

@Article{Ashcraft2026,
  author  = {Ashcraft, K.~R. and Grabowski, A.~M.},
  title   = {The effects of leg prosthesis stiffness and take-off board stiffness on long jump performance},
  journal = {Scientific Reports},
  year    = {2026},
  volume  = {16},
  pages   = {7418},
  doi     = {10.1038/s41598-026-38100-x}
}

@InBook{Arampatzis1999,
  chapter   = {Long jump},
  title     = {Biomechanical Research Project Athens 1997 Final Report},
  publisher = {Meyer \& Meyer Sport},
  year      = {1999},
  author    = {Arampatzis, A. and Brüggemann, G.-P. and Walsch, M.},
  editor    = {Brüggemann, G.-P. and Koszewski, D. and Müller, H.},
  address   = {Oxford},
}

@article{Hase2017,
author = {Hase, K. and Kobayashi, S. and Obinata, G. and Pei, Y.},
year = {2017},
month = {01},
pages = {D-32},
title = {Simulation and Optimization of the Takeoff Action and Prosthesis for Amputee Long Jump},
volume = {2017},
journal = {The Proceedings of the Symposium on sports and human dynamics},
doi = {10.1299/jsmeshd.2017.D-32}
}

\section*{Acknowledgements}

We want to thank the Simulation and Optimization research group of the IWR at Heidelberg University for giving us the possibility to work with MUSCOD-II. We want to thank Alena Grabowski, Hiroaki Hobara, Steffen Willwacher and Ralf Böhle for their help during the long jump project and data capturing.

\section*{Author contributions statement}
AE and KM prepared the models, the motion description and the optimal control problem. AE computed and analyzed the results.
JF and WP provided the motion capure recordings and biomechanical expertise. All authors discussed the results.
AE wrote the first draft of the manuscript, and the other authors reviewed it and provided detailed feedback. 

\section*{Funding declaration}
This research did not receive funding.

\section*{Additional information}

The authors declare no competing interests.

\end{document}